\documentclass[10pt,twocolumn,letterpaper]{article}
\usepackage{caption}

\usepackage{wacv}
\usepackage{times}
\usepackage{epsfig}
\usepackage{graphicx}
\usepackage{amsmath}
\usepackage{amssymb}
\usepackage{booktabs}

\def\wacvPaperID{1824} 

 \wacvfinalcopy 

\ifwacvfinal
\usepackage[breaklinks=true,bookmarks=false]{hyperref}
\else
\usepackage[pagebackref=true,breaklinks=true,colorlinks,bookmarks=false]{hyperref}
\fi

\begin{document}

\title{Anisotropic Multi-Scale Graph Convolutional Network for Dense Shape Correspondence}

\author{Mohammad Farazi, Wenhui Zhu, Zhangsihao Yang, and Yalin Wang\\
Arizona State University\\
Tempe, Arizona\\
{\tt\small \{mfarazi,wzhu59,zyang195,ylwang\}@asu.edu}
}

\maketitle
\thispagestyle{empty}

    
    

  
\begin{abstract}
This paper studies 3D dense shape correspondence, a key shape analysis application in computer vision and graphics. We introduce a novel hybrid geometric deep learning-based model that learns geometrically meaningful and discretization-independent features. The proposed framework has a U-Net model as the primary node feature extractor, followed by a successive spectral-based graph convolutional network. To create a diverse set of filters, we use anisotropic wavelet basis filters, being sensitive to both different directions and band-passes. This filter set overcomes the common over-smoothing behavior of conventional graph neural networks. To further improve the model's performance, we add a function that perturbs the feature maps in the last layer ahead of fully connected layers, forcing the network to learn more discriminative features overall. The resulting correspondence maps show state-of-the-art performance on the benchmark datasets based on average geodesic errors and superior robustness to discretization in 3D meshes. Our approach provides new insights and practical solutions to the dense shape correspondence research.
\end{abstract}

\section{Introduction}
In recent years, learning-based non-rigid shape correspondence approaches have been revolutionized with the success of geometric deep learning on unstructured data \cite{bronstein2017geometric}. These approaches come with various learning scenarios, from unsupervised to supervised learning methods on 3D mesh or point cloud representations \cite{halimi2019unsupervised,eisenberger2020deep,fey2018splinecnn,monti2017geometric,zeng2021corrnet3d}. While modeling dense shape correspondence problem can differ in terms of loss functions and problem formulations, the learned descriptors, ideally, have to be invariant to near-isometric deformations and discretization of 3D shapes. Unfortunately, many spatial-based graph neural network models, relying on neighbor feature aggregation, fail  to generalize to meshes with different sizes and discretizations as they overfit to the mesh connectivity \cite{sharp2022diffusionnet,wang2020mgcn,donati2020deep}.

\begin{figure}
  \centering
 \vspace{0.4cm}
  {\includegraphics[scale = 0.34, trim={1.8cm  1.5cm 3cm  0.8cm },clip]{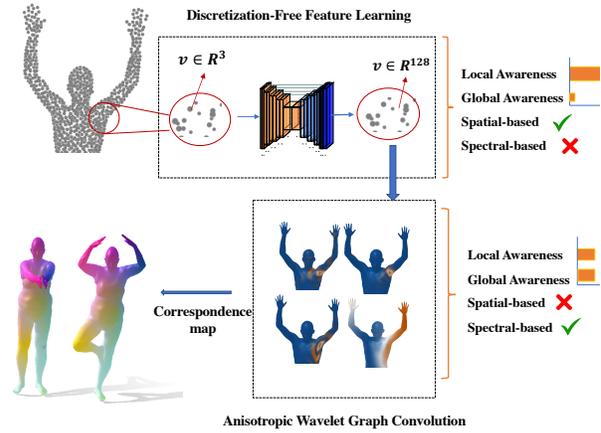}}
  
  \caption{Our hybrid model learns discretization-free features using a U-Net structure based on a spatial-based point cloud model, KPconv. \cite{thomas2019kpconv}. Then, to learn more globally and locally aware features, we use anisotropic wavelet based filters for our graph convolution layer to capture intrinsic information among different band-passes. Anisotropic wavelets also help our filters to be sensitive to direction.}  \label{fig:overal}
\end{figure}
Lately, studies in \cite{wang2020mgcn,donati2022deep,sharp2022diffusionnet,donati2020deep} have addressed the discretization invariance problem. The core to their approach is the idea of using the Laplace Beltrami Operator (LBO) eigen-basis and heat diffusion. The omnipresence of LBO can be used in many forms like the pre-computation of 3D shape wavelet filters \cite{wang2020mgcn},  heat diffusion as a spatial communication among the features ~\cite{sharp2022diffusionnet}, or general functional map settings ~\cite{ovsjanikov2012functional}. However, unlike state-of-the-art graph convolutional networks~\cite{li2020shape,wang2019dynamic,defferrard2016convolutional,velivckovic2017graph,fey2018splinecnn}, spectral models usually have inferior performance in single-resolution mesh setting in terms of perfect matched points and average geodesic errors. This can be seen as a trade-off between local and global feature learning awareness. To remedy this, introducing a more diverse set of filters in different band-passes and hybridization with spatial-based models can be a promising solution.

In this study, we propose  a hybrid model (Figure~\ref{fig:overal}) that incorporates the merits of both spatial and spectral-based approaches to overcome the limitations mentioned above. Our model learns in an end-to-end setting, starting with a point cloud U-Net block to learn geometric features and feed them to a spectral-based graph convolutional network to learn robust and discriminative features. To overcome the over-smoothing nature of graph convolutional networks, we adopt graph wavelet kernel as diverse filters to capture intrinsic information from different band-pass lenses. To make our filters even more diverse, we also propose to use the anisotropic LBO to make our kernels directionally sensitive. Finally, as the last layer of our network, we creatively use a layer to perturb the feature map to force the network to learn more discriminative features. As a result, it significantly improves our system performance.

Overall, our  contributions are summarized as follows: 
\newline\textbf{(1)} To our knowledge, it is the first  geometric deep learning framework that learns both spatial-based (via U-net) and spectral-based (via anisotropic wavelet graph convolution network) geometric features in a data-driven fashion. The learned  features are sensitive to subtle geometric changes and robust to 3D discretization differences.
\newline\textbf{(2)} We employ anisotropic wavelet filters to address the common over-smoothing drawback of conventional graph convolutional networks. Instead of only low-pass filtering, we design graph wavelet functions with different band-passes. Our work achieves a diverse set of filters sensitive to various directions and band-passes and effectively learns rich intrinsic geometric features.
\newline\textbf{(3)} We creatively apply a simple feature perturbation function in our last layer. It significantly boosts our model's performance in both average geodesic errors and convergence speed. The remarkable result may enrich our understanding of geometric learning strategy designs and inspire new architectures in the geometric deep learning field.

Our extensive experimental results verify the effectiveness of our method. We hope this work contributes to shape correspondence research and sheds new light on general geometric deep learning mechanisms to maximize their learning power.

\section{Related Works}

Shape correspondence computation methods generally fall into three main categories: $(1)$ traditional hand-crafted descriptors defined in spatial and spectral domains, $(2)$ optimization-based methods rooted in the spectrum of shapes with functional maps being the pillar, and $(3)$, more recent geometric deep learning techniques, notably graph neural networks. 

Starting with hand-crafted descriptors, they mainly fall into spatial and spectral methods. The spatial-based descriptors are usually based on statistics of locally defined features \cite{salti2014shot}. These methods, mostly, suffer from generalizability regarding changes in surface discretization and capturing global information of the shape \cite{wang2020mgcn}. This behavior is similar to spatial-based graph neural network frameworks that rely on learning small receptive fields around each vertex, making feature learning less globally aware. 
Spectral-based methods heavily rely on the LBO spectrum. Importantly, these intrinsic descriptors are isometry-invariant, making them robust to arbitrary spatial transformations. The methods proposed in \cite{aubry2011wave,bronstein2010scale,chaudhari2014global} belong to the spectral-based category. Most recently, wavelet-based spectral-based descriptors have been proposed in \cite{wang2020mgcn,li2021anisotropic}, exploiting a multi-scale setting to diversify the set of learned filters. However, our study does not rely on hand-crafted features as they are sub-optimal in learning-based models.


The second category of shape correspondence techniques is based on optimizing a dense map between shapes. Particularly, the vanguard of such methods is the successful functional map framework \cite{ovsjanikov2012functional}. More recently, the authors in \cite{donati2020deep,donati2022deep,litany2017deep} proposed unsupervised and supervised learning schemes on top of the functional map for a more robust and accurate correspondence. In \cite{donati2020deep}, authors proposed a U-Net structure to learn point-wise features to circumvent the pre-computation of hand-crafted features resulting in geometrically meaningful features independent of meshing structure. In \cite{litany2017deep}, authors used a deep residual network model to enhance the functional maps in creating a soft correspondence map. One of the most recent works on using a functional map based on learned features is the work in \cite{yang2021continuous}. They employed a continuous geodesic convolution in an end-to-end fashion. In another study~\cite{donati2022deep} with functional map setting, authors introduced an unsupervised framework for learning orientation-preserving features for functional map computation, which is also robust to discretization changes in the mesh. 
\begin{figure*}
 \centering
 \includegraphics[scale = 0.65,trim={2.0cm  5cm 1.0cm  5.2cm } ,clip]{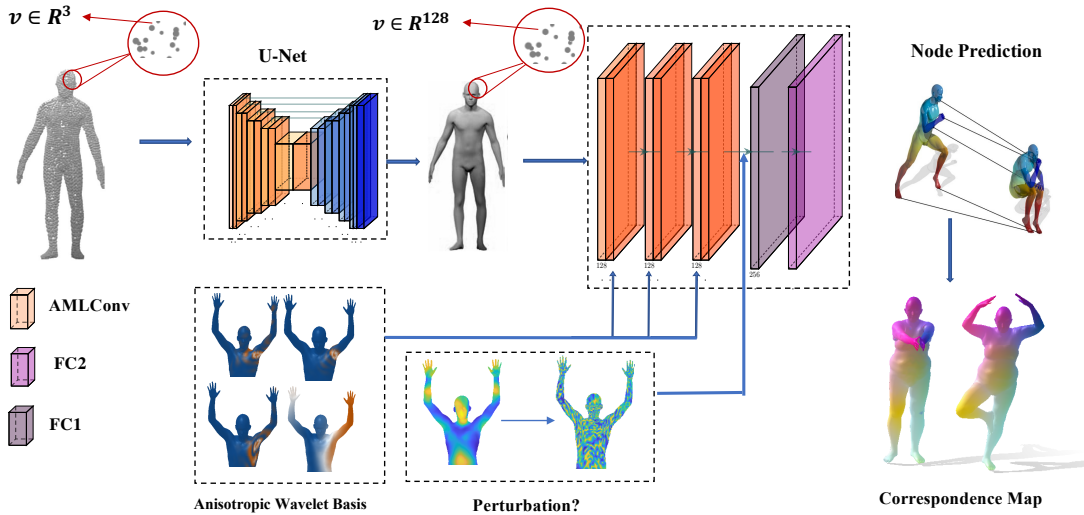}
 \caption{Our hybrid model at a glance. We use a U-Net feature extractor to learn geometric features from the raw point cloud; then we feed the geometric learned features and pre-computed wavelets $\Psi_{t_{\alpha\theta,j}}$ to our proposed anisotropic multi-scale graph convolution layers. Our network, therefore, learns local and global features, which are also robust to discretization. In the last layer, after we add a perturbation layer, the network forces to learn more discriminative features, as illustrated qualitatively in Figure ~\ref{fig:featmap}.}
 \label{fig:model}
\end{figure*}
The last category belongs to the state-of-the-art methods rooted in geometric deep learning, a new realm of deep learning in unstructured data like point clouds and meshes. In \cite{boscaini2015learning}, authors proposed windowed Fourier transform in a supervised-learning setting to learn local shape descriptors for shape matching problems Later, they used anisotropic LBO to learn multi-kernel filters to further improve the accuracy and diversity of learned features~\cite{li2021anisotropic}. Though mainly used for node and graph classification settings, graph neural network models have also been used for 3D dense shape correspondence. Frameworks in \cite{fey2018splinecnn,monti2017geometric,bouritsas2019neural} are among  methods that hold state-of-the-art performance in graph classification, 3D shape segmentation, and shape matching. Another model based on anisotropic LBO is proposed in \cite{li2020shape} and their experiment shows a boost in performance among other graph neural network-based models. Also, they showed with an explicit ablation study  how the different number of anisotropic kernels, i.e., diffusion kernels in different angles, affect the model's performance. In a different approach, the authors in \cite{guo2020learning} used a deep learning model to create local geometry images to learn descriptors based on local patches. Recently, \cite{sharp2022diffusionnet} proposed a geometric deep learning model using the concept of heat diffusion for spatial communication among the mesh equipped with spatial gradient features for better robustness. This method was applied to many shape analysis applications like segmentation, shape correspondence, and classification. However, an integration of spatial and spectral-based feature learning in the deep learning framework has not been fully studied. Our work aims to fill the gap by exploring this direction.

\section{Overview}
The overview of our hybrid system is illustrated in Figure~\ref{fig:model}.
We first introduce the geometrical foundations of building our anisotropic wavelet kernels to construct our graph convolution layers named \textbf{\underline{A}}nisotropic \textbf{\underline{M}}ulti-Sca\textbf{\underline{l}}e Graph \textbf{\underline{Conv}}olution (AMLCONV). Then, we will delve into the initial feature extractor block using Kernel Point Convolution (KPConv) \cite{thomas2019kpconv}, which is responsible for capturing geometry-aware features from the raw geometry input. Lastly, we will introduce a perturbation trick to improve the learning power in identifying  discriminative features.

\subsection{Geometrical background}
We model a 3D shape surface as a compact two-Riemannian manifold \textit{X} with set of vertices and edges as $(V, E)$, where $V = \{v_i|i=1,...,N\}$ and $E = \{e_ij|i,j=1,...,N\}$. We use $T_xX$ for each $x \in X$ to show the tangent plane of an arbitrary point on our shape $X$. The Riemannian metric is an inner product on this plane with the notation $\langle.,.\rangle_{T_{{x}_X}}: T_{{x}_X} \times T_{{x}_X} \longrightarrow R$, denoting how manifold locally deviates from the plane. To calculate this deviation and bending from the plane, the second fundamental form, represented by a 2*2 matrix, is used, whose eigenvalues are called principal curvatures $k_m$ and $k_M$. The corresponding eigenvectors form an orthogonal basis for $T_X$ on $x$, denoting the directions of curvatures. 
Many shape analysis applications rely on the LBO, on account of the local geometry based on the tangential plane, alleviating the problem by converting the calculations from 3D to 2D. Specifically, in this study, as introduced in \cite{andreux2014anisotropic}, we focus on the anisotropic version of LBO, namely, ALBO. 

\subsection{Anisotropic Laplace Beltrami Operator}
Before delving into ALBO, we define the omnipresent LBO. From a signal processing perspective, the eigen-system of LBO acts as Fourier bases on signals. However, on a 2-Manifold, the bases are not periodic functions like sine and cosine. Although orthogonal functions like spherical harmonics have the same property as pseudo-spherical shapes, the LBO eigen-system is generally used for geometry processing applications on deformed shapes. The definition of LBO on shape $X$ is
\begin{equation} \label{eq1}
\Delta_X f(x) = -div_X(\nabla_X f(x))
\end{equation}
where $\Delta_X$ and $div_X(.)$ are, respectively, intrinsic gradient and divergence of $f(x) \in L^2(x)$. Due to the positive semi-definiteness of the LBO operator, we have the real eigen-decomposition 
\begin{equation} \label{eq2}
\Delta_X \phi_k(x) = \lambda_k \phi_k(x),
\end{equation}
with non-negative eigen-values $\{\lambda_i \leq \lambda_j| i < j\}$ and eigen-functions $\{\phi_i(x)\}$ forming an orthonormal basis for $L^2(x)$. Due to the highly expensive eigen decomposition of LBO, as it has been used in numerous applications \cite{ovsjanikov2012functional,litany2017deep,wang2020mgcn}, we use the first $k$ eigen-values and eigen-functions to circumvent the curse of costly computation. 

The LBO is both intrinsic and isotropic, meaning it is invariant under isometric deformation and is not sensitive to directions. However, by making it dependent on direction $\theta$, we can exploit its sensitiveness among different deformations in different directions. In other words, we try to embed extrinsic geometry into intrinsic by introducing ALBO. Now, the new equation is 
\begin{equation} \label{eq3}
\Delta_X f(x) = -div_X(D(x)\nabla_X f(x)),
\end{equation}
where $D$ is a 2*2 matrix named anisotropic tensor or thermal conductivity tensor acting on the intrinsic gradient. By introducing $D$ the diffusion can be controlled in both direction and magnitude but to have a standard direction among all points, researchers suggested anisotropy along maximum curvature resulting in a matrix 
\begin{equation} \label{eq4}
D_\alpha(x) = 
\begin{pmatrix}
  \frac{1}{1+\alpha} & \\
   & 1
\end{pmatrix},
\end{equation} 
where $\alpha$ controls the anisotropy level, instead of considering one direction, we can consider multiple directions for our kernel \cite{andreux2014anisotropic}. The definition of new anisotropic tensor $D$ is as follows
 \begin{equation} \label{eq5}
D_{\alpha\theta}(x) = R_{\theta}
\begin{pmatrix}
  \frac{1}{1+\alpha} & \\
   & 1
\end{pmatrix}
R_{\theta}^T,
\end{equation}
By setting the reference angle $\theta = 0$ to principal curvature direction, we can form a set of rotations in $[0,2\pi]$. Now, the eigen decomposition of the new LBO in Eq.~\ref{eq2} becomes
\begin{equation} \label{eq6}
\Delta_{\alpha\theta} \phi_{\alpha\theta,k}(x) = \lambda_{\alpha\theta,k} \phi_{\alpha\theta,k}(x) ,
\end{equation}
where our ALBO still satisfies the semi-positive definiteness and forms a set of orthogonal eigen-functions $\phi_{\alpha\theta,k}(x)$. As mentioned for the LBO, here for ALBO we compute the first $k$ eigen-values and eigen-functions.
\begin{figure*}
 \centering
 \includegraphics[scale = 0.53,trim={1.0cm  4.8cm 1.5cm  4cm },clip]{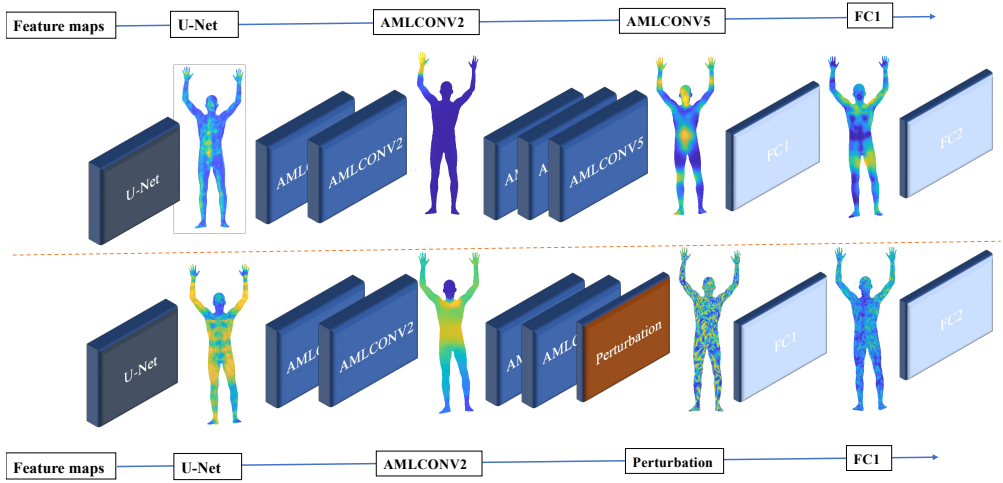}
 \caption{Intermediate feature maps comparison between our vanilla and perturbed version model. Top: Network with no perturbation layer. Bottom: Network with a perturbed feature map as the last layer before fully connected layers. As shown in the perturbed version, the feature maps of previous layers become more semantically meaningful and distinctive. }
 \label{fig:featmap}
\end{figure*}
\subsection{3D Mesh Wavelets}

A graph wavelet function is locally defined on vertex $v$ denoted as $\Psi_{t,v}$ per scale $t$. Wavelet Graph Neural Network models employed in \cite{chang2021spectral,xu2019graph} use just one scaling function instead of considering all the scales for $t$, which negates the spirit of multi-scale analysis \cite{xu2019graph,chang2021spectral}. We need a kernel function $g$ to sweep the frequencies in the spectral domain for different band-passes. Mexican Hat and Meyer wavelets are among the many wavelet functions that frequently appear in the literature. Unlike graphs, we deal with a discretized two-manifold with a metric, and Graph Laplacian no longer captures the geometry of the mesh. We rather use the introduced ALBO in the previous section. Using the area matrix $A$ for $X$, we define the Voronoi area $a(v)$ at vertex $v$, $\phi_j(v)$ the eigen-function corresponding to vertex $v$ to define the localized graph wavelet at vertex $v$
\begin{equation} \label{eq7}
\Psi_{\alpha\theta, t,v} = \sum_{j=0}^{N-1} a(v)g(t\lambda_{\alpha\theta,j})\phi_{\alpha\theta,j}(v)\phi_{\alpha\theta,j}.  
\end{equation}
The low pass filter representing the wavelet basis is called the scaling function and is defined based on filter $h(x)$.
\begin{equation} \label{eq8}
\xi_{\alpha\theta,v} = \sum_{j=0}^{N-1} a(v)h(\lambda_{\alpha\theta,j})\phi_{\alpha\theta,j}(v)\phi_{\alpha\theta,j}.  
\end{equation}

To find the wavelet coefficients for both scaling and wavelet functions, we have to project the given function $f$ onto the wavelet basis. Coefficients are for spectral wavelet, and its scaling function is derived by
\begin{equation} \label{eq9}
W_f(\alpha\theta,t,v) = \sum_{j=0}^{N-1} a(v)g(t\lambda_{\alpha\theta,j})\sigma_{\alpha\theta,j}\phi_{\alpha\theta,j}(v),  
\end{equation}
\begin{equation} \label{eq10}
S_f(\alpha\theta,v) = \sum_{j=0}^{N-1} a(v)h(\lambda_{\alpha\theta,j})\sigma_{\alpha\theta,j}\phi_{\alpha\theta,j}(v).  
\end{equation}
If $h$ and $g$ are chosen to be a Parseval frame, we can reconstruct the signal using the equation below \cite{wang2020mgcn}:

\begin{equation} \label{eq11}
f= \sum_{j=0}^{K} \sum_{v} a(v)^{-1}W_f(\alpha\theta,t_j,v)\Psi_{\alpha\theta,t_j,v}. 
\end{equation}

\subsection{Feature Extractor}
This block is responsible for learning the geometric characteristics of the mesh as the input for the AMLCONV layers. 
We use the same point cloud learning method KPConv \cite{thomas2019kpconv} with a U-net architecture that was primarily used for the segmentation task in the main study. Using the learned features helps us with more informative and data-centric features to build our multi-scale network on top of it. 

    


Unlike the previous works \cite{wang2020mgcn,ovsjanikov2012functional} using hand-crafted features, the input to the feature extractor network is a point cloud with just the $3D$ coordinates as input. The U-Net architecture is composed of 4 down-sampling and four up-sampling layers equipped with convolutional layers, i.e., KPConv. For more information on the kernel used in KPConv, we refer readers to~\cite{thomas2019kpconv}. As for pooling, grid sampling is used for flexibility over density control at each pooling layer.

\subsection{Convolutional Layer Setup}

The feature extractor block outputs $D'$ dimensional learned features  to the main multi-scale layer, which is a graph convolutional neural network architecture. Therefore, considering the input signal $x$ to our multi-scale layer $\in R^{N \times D'}$ indicating $N$ vertices and $D'$ dimensional signal on each point. The multi-scale layer, as witnessed in \cite{defferrard2016convolutional}, convolves the signal $x$ with filter $g$ in the spectral domain as follows
\begin{equation} \label{eq12}
x*g = \Phi((\Phi^Tg)\odot(\Phi^Tx)) = \Phi W\Phi^Tx,
\end{equation}
where $\odot$ is the element-wise product and $\Phi$ is the $k$ first eigen vector matrix $\Phi \in R^{N \times k}$, and $w$ is the filter represented in the frequency domain. Mainly the focus of studies is on how to design the filter $w$. In ChebyNet \cite{defferrard2016convolutional}, weighted sum of powers of eigen-value matrix is used, circumventing the use of eigen decomposition. Specifically, they used polynomial powers of eigen-value matrix $\Lambda$ which can be computed recursively and induce $k$ hop receptive field kernel in spatial domain with approximation order $k$
\begin{equation} \label{eq13}
w = \sum_{j=0}^{k-1} \theta_j (\Lambda)^j,
\end{equation}
where the diagonal matrix $\Lambda$ is of size $k*k$, and the convolution equation in Eq.~\ref{eq2} becomes 
\begin{equation} \label{eq14}
x_{out} = g * x = \sum_{j=0}^{k-1} \theta_j T_j(L)x,
\end{equation}
where $\theta_j$ are the set of learnable parameters in the network and $T_j(L)$ is the $j$ order Chebyshev Polynomials. As a result, we have an m-localized receptive filed around each vertex $v$. There are two drawbacks to such a filter. One is that this filter is an all-pass filter, and secondly, it is not resolution independent as it overfits to the mesh connectivity.

Here we use $w$ in terms of wavelet filter basis in the spectral domain rather than any polynomials mentioned above. This is beneficial to track different frequency band-passes and not just use a low-pass filter resulting  in over-smoothness of the signals on the vertices
\begin{equation} \label{eq15}
W = \sum_{j=0}^{k} \theta_j f_{t_j}(\Lambda),
\end{equation}
and the corresponding convolution is
\begin{equation} \label{eq16}
x_{out} = g * x =\sum_{j=0}^{k} \theta_j \Psi_{t_j}^Tx,
\end{equation}
where the wavelet basis matrices at each scale $t_j$, $\Psi_{t_j} \in R^{N \times N}$, are used for the convolution. Since the magnitude of different scales of $\Psi$ varies, we use normalization for a better learning process. We perform $L_1$ normalization on the wavelet functions. We show the normalized wavelet basis functions as $\overline{\Psi}$, and the new convolution becomes \cite{wang2020mgcn}
\begin{equation} \label{eq17}
x_{out} = g * x =\sum_{j=0}^{k} \theta_j \overline{\Psi_{t_j}}^Tx.
\end{equation}
Consequently, the multi-scale layer, including a custom batch normalization ``Norm" and activation function ``SELU" with isotropic LBO, is as follows
\begin{equation} \label{eq_iso}
\text{Norm}\left(\text{SELU}\left(\sum_{j=0}^{k} \theta_j \overline{\Psi_{t_j}}^Tx\right)\right).
\end{equation}

Finally, the anisotropic multi-scale convolution layer based on the definition of anisotropic LBO in 1.5 and 1.6 by setting $\alpha$ as a hyper-parameter of our network is
\begin{equation} \label{eq_aniso_eq}
\text{Norm}\left(\text{SELU}\left(\sum_{\theta} \sum_{j=0}^{k} \theta_{\alpha\theta,j} \overline{\Psi_{t_{\alpha\theta,j}}}^Tx\right)\right).
\end{equation}

\begin{figure}[t]
  \centering
  
  {\includegraphics[scale = 0.35,trim={1cm  3.5cm 0cm  3.8cm },clip]{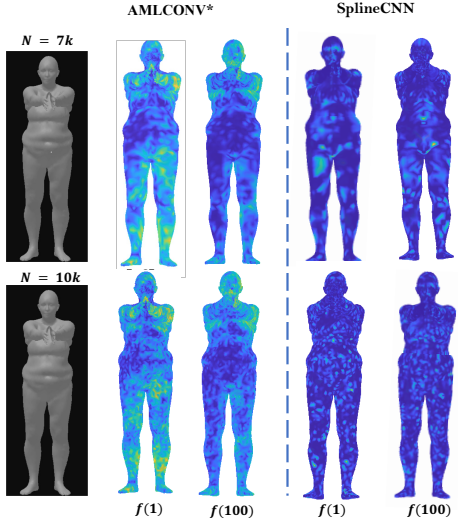}}
  
  \caption{Feature map ($f$) consistency among two different resolutions of the same shape. $f(x)$ shows the $x-$ dimension of feature map. As illustrated, features learned by our model are semantically and visually more consistent in both resolutions, whereas in splineCNN, there is no consistency and smoothness}.
  \label{fig:faustVsSpline_plot}
\end{figure}

\begin{table}[t]
 \caption{Average geodesic error on all pairs of ($15 \times 14$) shapes in FAUST test dataset. The results, for the sake of fair comparison, excluding ours, are derived from \cite{wang2020mgcn,sharp2022diffusionnet}. All average geodesic errors are reported $\times 100$. In the table,``Ours" refer to our vanilla model, and ``ours*" refer to our model using perturbation layer. Terms WEDS and XYZ refer to ``Wavelet Energy Decomposition Signature" ~\cite{wang2020mgcn}, and ``3D xyz" coordinate of vertices. }\label{tab1}
\begin{tabular}{ |p{3.1cm}|p{1cm}|p{1cm}|p{1cm}| }

\hline
\multicolumn{1}{|c|}{Network} &\multicolumn{1}{|c|}{Descriptor} & \multicolumn{2}{c|}{Resolution} \\
\cline{3-4}
\multicolumn{1}{|c|}{} &\multicolumn{1}{|c|}{} & 7k & 10k  \\
\hline
{OSD \cite{litman2013learning}} &  1 & 39.8 & 45.7  \\\hline
{SplineCNN \cite{fey2018splinecnn}} & 1 & 27.6 & 52.4  \\\hline
{ChebyGCN \cite{wang2020mgcn}}    &WEDS & 0.6& 55.1 \\ \hline
{Geo-Based \cite{wang2019robust}}    &WEDS &14.7& 23.9 \\ \hline
{DiffusionNet \cite{sharp2022diffusionnet}}    &XYZ &0.33& \textbf{0.83} \\ \hline
{ACSCNN \cite{sharp2022diffusionnet,li2020shape}}    &1 & \textbf{0.05}& 41.1 \\ \hline
{MGCN \cite{wang2020mgcn}}    &WEDS &0.8& 2.6 \\ \hline
{U-Net (KPconv \cite{thomas2019kpconv}}    &XYZ &2.8& 5.7 \\ \hline
{\textbf{Ours}}    &XYZ &\textbf{0.8} & \textbf{2.4} \\ \hline
{\textbf{Ours*}}    & XYZ & \textbf{0.12} &  \textbf{0.85} \\ 
\hline

\hline

\end{tabular}
\end{table}

\begin{table}[t]
 \caption{Average geodesic error on all test pairs ($10 \times 9$) of shapes in SCAPE test dataset. The results, for the sake of fair comparison, excluding ours are derived from \cite{wang2020mgcn,sharp2022diffusionnet}. All average geodesic errors are reported $\times 100$. In the table,``Ours" refer to our vanilla model, and ``ours*" refer to our model using perturbation layer.}\label{tab2}
\begin{tabular}{ |p{3.1cm}|p{1cm}|p{1cm}| }

\hline
\multicolumn{1}{|c|}{Network} &\multicolumn{1}{|c|}{Descriptor} & \multicolumn{1}{c|}{Resolution (5k)} \\
\hline
{OSD \cite{litman2013learning}} &  1 & 25.9   \\\hline
{SplineCNN \cite{wang2020mgcn,fey2018splinecnn}} & 1 & 29.7   \\\hline
{ChebyGCN \cite{wang2020mgcn}}    &WEDS & 6.8 \\ \hline
{Geo-Based \cite{wang2020mgcn}}    &WEDS &16.3 \\ \hline
{DiffusionNet \cite{sharp2022diffusionnet}}    &XYZ &\textbf{3.0} \\ \hline
{MGCN \cite{wang2020mgcn}}    &WEDS &4.8 \\ \hline
{\textbf{Ours}}    &XYZ &\textbf{3.4}  \\ \hline
{\textbf{Ours*}}    & XYZ & \textbf{1.2}    \\ 
\hline

\hline

\end{tabular}
\end{table}

\subsection{A Perturbation Layer}
In this study, we discovered that using a perturbation layer at the end of the network can accelerate training convergence and performance by a large margin. For the last layer, instead of using Eq.~\ref{eq19}, we can use a fixed shuffling of features learned in the previous layer $x_{n-1}$. This permutation of features acts as a perturbation of the whole feature map. Surprisingly, this forces the entire network, including the primary U-Net block, to learn more discriminative and robust features. The formulation of one possible perturbation is as follows:
\begin{equation} \label{eq19}
\text{Norm}\left(\text{SELU}\left( \theta * x_{perm}\right)\right),
\end{equation}
The perturbation term in the above equation must be fixed for all batches. For clarification, if this perturbation changes the node feature $i$ and $j$ for the first batch, it must be done for the remaining batches. Note that we still have learnable parameters $\theta$ which control this perturbation's weight for each feature. We illustrate a shape's intermediate feature maps after training on both training schemes, with and without perturbation, in Figure~\ref{fig:featmap}. Although the final feature map of the perturbed network is less smooth, it clearly shows how this perturbation results in a semantically discriminative feature map in previous layers. In our study, we simply use a matrix transpose of feature maps to shuffle the features among nodes.  
\subsection{Network Details}
The KPconv U-Net network is a residual-based point cloud convolutional network, with four convolutional layers with leaky linear activation functions and poolings and successive up-sampling layers to the final $128$ feature dimension. The step-size of sub-sampling is set to $0.3$ as defined in \cite{donati2020deep}. Next, we use four layers of AMLCONV with input and out feature dimension of $128$, with a fully connected layer and a softmax at the end to predict the labels of each vertex for shape matching. We use scaled exponential linear units ``SELU" activation function for our AMLCONV layers. 

As for the wavelet function candidate, we use the Mexican Hat filter following the same setting in \cite{wang2020mgcn}. The function $g$ is as follows:
\begin{equation} \label{eq_mexican}
x_{out} = g * x =\sum_{j=0}^{k} \theta_j \Psi_{t_j}^Tx,
\end{equation}

\begin{figure}[t]
  \centering
  \hspace*{-0.4cm}
  {\includegraphics[trim={3cm  2cm 4cm  1.3cm },clip,scale = 0.35]{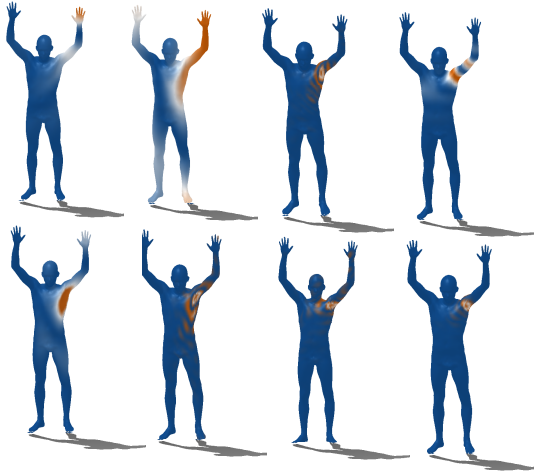}}
  
  \caption{Visualization of different anisotropic wavelet filters on the same mesh with the same selected node on the right hand. A sequence of results from the low-pass, band-pass, and high-pass filters with different anisotropy levels are illustrated. It is conceivable from these kernel maps that wavelet filters act both locally and globally for robust feature learning.}
  \label{fig:filters_plot}
\end{figure}

We use $16$ different filters, including $4$ different anisotropy directions, and for each, we compute 4 different wavelet functions. For the LBO eigen-basis, we computed the first $200$ eigen-values and eigen-vectors to compute our wavelet filters accordingly. In Figure~\ref{fig:filters_plot} different filters, from the low-pass to the high-pass, and different diffusivity directions are illustrated for a specific node on a shape.

  
\subsection{Experimental Results}
\textbf{Datasets}. To test our network, we apply it to two publicly available datasets, FAUST \cite{bogo2014faust} and SCAPE, \cite{anguelov2005scape} to evaluate our descriptor learning performance for dense shape correspondence in \textit{near-isometric} shapes. All the datasets are provided with ground-truth labels for shape correspondence and consist of deformable shapes of human poses. We also use another re-meshed version of FAUST \cite{wang2019robust,wang2020mgcn} to evaluate the robustness of our model in terms of re-meshing and variability in size. The re-meshed FAUST dataset maintains the coordinates of the original $7k$ resolution dataset. The resolution we choose is $10k$ for testing on the trained model with $7k$ resolution. A total of 15 shapes are used for the test set in FAUST dataset. For the SCAPE dataset, we only use the simplified re-meshed version based on the setting in \cite{yan2014low} containing approximately $5k$ vertices. This re-meshing, however, changes the position and triangulation \cite{wang2020mgcn}. For the sake of fair comparison, we report the results based on the main benchmark study in \cite{wang2020mgcn}. In SCAPE dataset, we used $10$ shapes for the test set.

\noindent\textbf{Evaluation criteria}. Per \cite{li2021anisotropic}, we take the Princeton benchmark protocol \cite{kim2011blended} to evaluate the correspondence accuracy based on the percentage of matches that are at most r$-$distant from the ground-truth map. As commonly used in the literature~\cite{wang2020mgcn}, we call it Cumulative Geodesic Error (CGE). The other criterion we use to measure the performance is the average geodesic errors in all test pairs. To find the predicted point in each test pair of the source-target shapes, we use nearest-neighbor searching using the L2 distance in the feature space.

\noindent\textbf{Experimental setup}
All experiments were done on a $V100$ Nvidia GPU, $32GB$ of RAM using Pytorch-Geometric packages \cite{fey2019fast} (Python $3.7$). The training was done in $200$ epochs for the vanilla network and $50$ epochs with our network with the perturbation layer. In both settings, the training was done with the ADAM optimizer \cite{kingma2014adam} with a learning rate of $0.001$ and weight decay of $0.0001$.  

\begin{figure}[t]
  \centering
  \hspace*{-1cm}
  {\includegraphics[scale = 0.35,trim={0cm  5cm 2.5cm  5.3cm },clip]{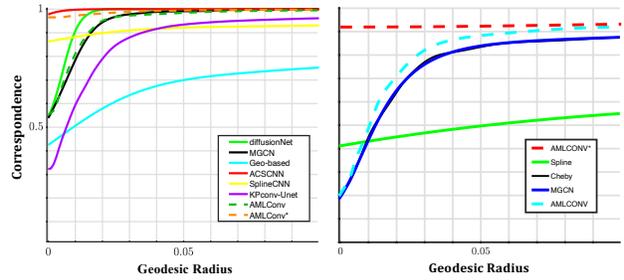}}
  
  \caption{Left: CGE of FAUST dataset. Right: CGE of SCAPE dataset. Dashed lines indicate the resolution of the test set with our models using both vanilla  and perturbed feature versions. As it is depicted, our perturbed version results in state-of-the-art outcomes in both datasets.}
  \label{fig:faust_plot}
\end{figure}

\begin{figure}[t]
  \centering
  \hspace*{-0.5cm}
    {\includegraphics[scale=0.23,trim={1.5cm  1cm 2.5cm  2.4cm },clip]{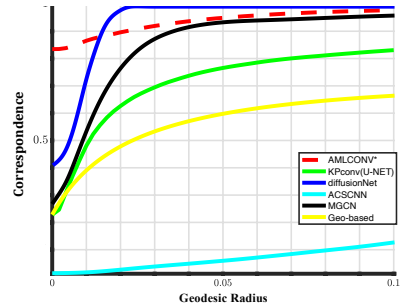}}
  
  \caption{CGE of FAUST dataset after training on original meshes and testing 10k re-meshed shapes. Our model still results in state-of-the-art CGE besides DiffusionNet. }
  \label{fig:10k_plot}
\end{figure}

\noindent\textbf{Results}
Quantitative results based on the average geodesic error on the FAUST dataset are tabulated in Table~\ref{tab1}. For the sake of fair comparison, the results with the baseline selected in \cite{wang2020mgcn} and \cite{sharp2022diffusionnet} are reported here. Graph neural network approaches that produce state-of-the-results in terms of minimum geodesic error like ACSCNN, ChebyNet, and SplineCNN show superior results in the same resolution. Our results using the perturbation layer beside ACSCNN have the best CGE. However, our model, unlike ACSCNN and SplineCNN, significantly outperforms other models, except DiffusionNet, after testing on the re-meshed dataset. DiffusionNet and our model are within the same margin of error when testing on different mesh sizes. In Figure \ref{fig:faust_plot} the CGE of different methods is depicted for the original FAUST dataset ($7k)$. The results after testing on $10k$ meshes are illustrated in Figure~\ref{fig:10k_plot}.  

To show the consistency of feature maps among different resolutions of shapes, in Figure~\ref{fig:faustVsSpline_plot}, we compare our results with SplineCNN. There is a high level of consistency among the feature maps in our model while tested on different shape resolutions. 

As for the SCAPE dataset, the quantitative results based on average geodesic error  are tabulated in Table~\ref{tab2}. The results show the superior performance of our model over other benchmarks. Even our vanilla network improves upon MGCN, the closest benchmark to ours. The vanilla network is also within the same margin of error as DiffusionNet in functional map setup. The CGE plots are depicted in Figure~\ref{fig:faust_plot}, showing the high number of perfect matching points based on our model with the perturbation layer. The convergence in our model with the perturbation layer is reached within $50$ epochs, making it $4$ times faster than the vanilla network and MGCN \cite{wang2020mgcn} with $200$ epochs with the same time complexity.

\noindent\textbf{Ablation study}
The important part of our model is the contribution of each part and parameters to the final learned descriptors. Since we use KPconv as the primary network block, we test a U-Net network followed by two fully connected layers to measure the performance of the U-Net feature extractor solely. Moreover, we test our model based on two training paradigms, with and without proposed perturbation, to see how feature maps would look like for better comparison. Lastly, since the number of wavelet filters would directly impact the learned descriptor, we also test our model based on a different number of anisotropic filters. 

\begin{figure}[t]
  \centering
  \hspace*{-0.4cm}
  {\includegraphics[scale=0.3,trim={2cm  3.4cm 2.5cm  3.5cm },clip]{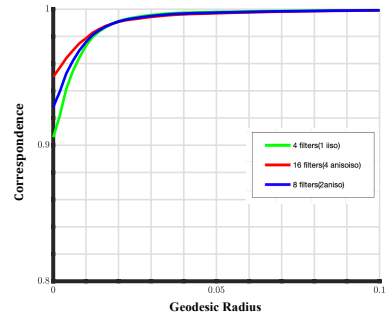}}
  
  \caption{Direct CGE in the FAUST dataset with a different number of filters. We test three different settings, one with $4$ wavelet filters using LBO. Second, with $8$ different filters, with anisotropy of degree two. In the third setting, we use all $16$ filters with $4$ anisotropy levels. Results clearly show that  more filters results in better performance. For more ablation results on number of filters please see ~\cite{li2021anisotropic,wang2020mgcn}.}
  \label{fig:ablation1_plot}
\end{figure}

We show our ablation study results in Figure ~\ref{fig:ablation1_plot}. As the number of filters increases, with different anisotropy levels and wavelet basis, the CGE improves. Using only the U-Net block for learning descriptor on FAUST dataset shows inferior results as opposed to our hybrid model. The average geodesic error is $2.8$ which is worse than the hybrid model.

\noindent\textbf{Limitations} In our study, although the perturbation layer improves the model's performance, it remains to be meticulously studied in the future for the potential theoretical reason behind this behavior. There can be more efficient perturbation other than the shuffling feature map that we did to reduce the geodesic error.



  

\section{Conclusion and Future Works}

In this study, we propose a hybrid end-to-end model taking advantage of the merits of spectral wavelet filters and geometry-aware spatial point cloud convolution model, KPconv, to overcome the overs-smoothing nature of most graph neural network models. Using the U-Net model, the learned geometric features are fed into our AMLCONV layers to learn robust descriptors using different band-pass filters. To further diversify the set of filters, we adopt the anisotropic LBO to capture directionally sensitive information. Later, we propose and depict how a simple perturbation of the feature map in the last layer of the network significantly improves the performance and convergence speed. 
In the future, we will focus on the theoretical foundation of feature perturbation and explore how it can be generalized to, possibly, other existing models.

\noindent\textbf{Acknowledgment}: The research is partly supported by NIH (R21AG065942, R01EY032125, R01EB025032, and R01DE030286).



{\small
\bibliographystyle{ieee_fullname}
\bibliography{egbib}
}

\end{document}